\documentclass[letterpaper, 10 pt, conference]{ieeeconf}  

\IEEEoverridecommandlockouts                              

\usepackage{tabu}
\usepackage{graphicx}
\usepackage{newclude}
\usepackage{color}
\usepackage[dvipsnames, svgnames, x11names]{xcolor}
\usepackage[colorlinks,linkcolor=blue,breaklinks=true]{hyperref}
\usepackage[para,online,flushleft]{threeparttable}
\usepackage{etoolbox}
\usepackage{stfloats}
\usepackage{subcaption}
\usepackage{amsmath}
\usepackage{ulem}
\usepackage{tikz}
\usepackage{cite}
\usepackage{url}
\usepackage{verbatim}
\usepackage{booktabs}
\usepackage{pifont}
\usepackage{algorithm}
\usepackage{algpseudocode}
\usepackage{dsfont}
\usepackage{amsfonts} 
\usepackage{setspace}
\usepackage[skip=1.5pt]{caption}
\usetikzlibrary{positioning}
\usetikzlibrary{matrix}
\usetikzlibrary{shapes.multipart} 
\usetikzlibrary{arrows.meta}
\usetikzlibrary{calc}
\makeatletter
\patchcmd{\@makecaption}
  {\scshape}
  {}
  {}
  {}
\makeatletter
\patchcmd{\@makecaption}
  {\\}
  {.\ }
  {}
  {}
\makeatother

\title{\LARGE \bf
Pick-and-place Manipulation Across Grippers Without Retraining:
\\ A Learning-optimization Diffusion Policy Approach
}

\author{Xiangtong Yao$^{1*}$, Yirui Zhou$^{1*}$, Yuan Meng$^{1}$, Liangyu Dong$^{1}$, Lin Hong$^{2}$,\\Zitao Zhang$^{1}$, Zhenshan Bing$^{1,3\dagger}$, Kai Huang$^{4}$, Fuchun Sun$^{5}$, Alois Knoll$^{1}$
\thanks{$^1$ Technical University of Munich, Munich, Germany}
\thanks{$^2$ Hong Kong University of Science and Technology, China}
\thanks{$^3$ Nanjing University, Nanjing, China}
\thanks{$^4$ Sun Yat-sen University, Guangzhou, China}
\thanks{$^5$ Tsinghua University, Beijing, China}
\thanks{$^*$ Equal contribution}
\thanks{$^{\dagger}$Corresponding author: Zhenshan Bing {\tt\small zhenshan.bing@tum.de}}}

\providecommand{\keywords}[1]{\textbf{\textit{Keywords---}} #1}

\let\oldtwocolumn\twocolumn
\renewcommand\twocolumn[1][]{%
    \oldtwocolumn[{#1}{
    \vspace{-5pt}
    \begin{center}
           \includegraphics[width=\textwidth]{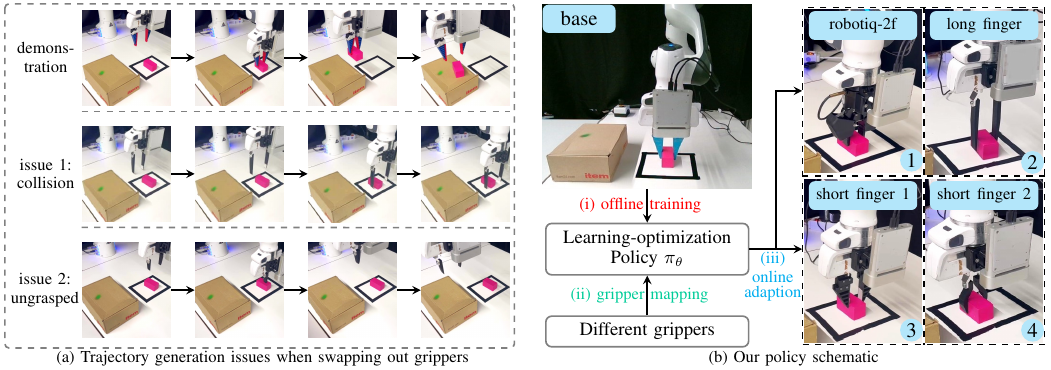}
           \captionof{figure}{(a) Different issues of the trajectory generation of the vanilla diffusion policy across grippers without the policy retaining. (b) Our policy aims to transfer pick-and-place skills across grippers without policy retraining while ensuring task completion, involving (i) offline training the policy $\pi_\theta$ with data collected via a base gripper, (ii) gripper-aware mapping, and (iii) online optimization adaption to enforce the generated trajectories to fit different unseen grippers.} 
           \label{fig:first-figure}
    \end{center}
    }]
}

\begin{document}

\maketitle

\begin{abstract}
Current robotic pick-and-place policies typically require consistent gripper configurations across training and inference. This constraint imposes high retraining or fine-tuning costs, especially for imitation learning-based approaches, when adapting to new end-effectors. To mitigate this issue, we present a diffusion-based policy with a hybrid learning-optimization framework, enabling zero-shot adaptation to novel grippers without additional data collection for retraining policy. During training, the policy learns manipulation primitives from demonstrations collected using a base gripper. At inference, a diffusion-based optimization strategy dynamically enforces kinematic and safety constraints, ensuring that generated trajectories align with the physical properties of unseen grippers. This is achieved through a constrained denoising procedure that adapts trajectories to gripper-specific parameters (e.g., tool-center-point offsets, jaw widths) while preserving collision avoidance and task feasibility. We validate our method on a Franka Panda robot across six gripper configurations, including 3D-printed fingertips, flexible silicone gripper, and Robotiq 2F-85 gripper. Our approach achieves a 93.3\% average task success rate across grippers (vs. 23.3-26.7\% for diffusion policy baselines), supporting tool-center-point variations of 16–23.5 cm and jaw widths of 7.5–11.5 cm. The results demonstrate that constrained diffusion enables robust cross-gripper manipulation while maintaining the sample efficiency of imitation learning, eliminating the need for gripper-specific retraining. Video and code are available at \href{https://github.com/yaoxt3/GADP}{https://github.com/yaoxt3/GADP}.

\end{abstract}

\definecolor{color-obs-seq}{RGB}{219, 232, 213}
\definecolor{color-act-seq}{RGB}{172, 204, 255}
\definecolor{color-opti}{RGB}{248,206,204} 
\definecolor{color-gmap}{RGB}{0,204,153} 

\definecolor{ffblue}{RGB}{097, 108, 140}
\definecolor{ffdarkgreen}{RGB}{086, 140, 135}
\definecolor{fflightgreen}{RGB}{178, 213, 155}
\definecolor{ffyellow}{RGB}{242, 222, 121}
\definecolor{ffred}{RGB}{217, 095, 024}
\definecolor{ffred_pv}{RGB}{202, 074, 046}
\definecolor{fforange_pv}{RGB}{232, 141, 047}
\definecolor{ffgreen_pv}{RGB}{059, 165, 149}
\definecolor{ffgreendark_pv}{RGB}{032, 117, 106}
\definecolor{nature_tab_gray1}{HTML}{D8D6C2}
\definecolor{nature_tab_gray2}{HTML}{ECEADF}

\definecolor{graspw}{RGB}{0, 128, 0}
\definecolor{dp}{RGB}{64, 224, 208}
\definecolor{dp3}{RGB}{63, 63, 255}
\definecolor{ours}{RGB}{148, 0, 211}
\definecolor{blockcolor}{RGB}{238, 130, 238}
\definecolor{safeline}{RGB}{255, 0, 0}
\definecolor{bananacolor}{RGB}{255, 165, 0}


\section{Introduction}

Robotic pick-and-place operations form the cornerstone of automated manipulation, which is essential for activities from household to industrial scenarios. A variety of methods have been proposed to endow robots with pick-and-place capabilities, including reinforcement learning~\cite{10341769,10160626,meng2025preserving}, imitation learning~\cite{10685120}, and motion planning~\cite{migimatsu2020object}. While these methods have demonstrated the ability to manipulate multiple objects (known or unknown)~\cite{9131812,zeng2021transporter}, operate in dynamic environments~\cite{migimatsu2020object}, or perform one-shot or few-shot tasks~\cite{di2023one,10202179}, they typically assume fixed end-effector morphology throughout training and deployment. This assumption restricts the transfer of pick-and-place skills across grippers and compromises overall generalizability.

Recent research mitigates this limitation through multi-embodiment learning strategies. For example, HPT trains a transformer-based policy with large-scale heterogeneous robot datasets covering diverse embodiments and tasks~\cite{wang2024scaling}. Wang et al.~\cite{wang2024cross} employ latent space alignment to transfer pick-and-place skills between different robot manipulators. Despite promising results, these methods often require substantial embodiment-specific data \cite{wang2024scaling} or policy fine-tuning to accommodate new gripper configurations \cite{wang2024scaling,wang2024cross}, leading to additional adaptation costs and limiting deployment flexibility. This challenge is particularly pronounced for behavior cloning algorithms, such as diffusion policies~\cite{chi2023diffusion,Ze2024DP3}. In online developments, switching from a base gripper to one with a different morphology can shift the end-effector’s tool-center-point (TCP), causing collisions or missed grasps if a pretrained diffusion policy is naively applied to generate trajectories, as shown in Fig. \ref{fig:first-figure}\textcolor{blue}{a}.

We bridge this gap through an integration of diffusion policy~\cite{chi2023diffusion,Ze2024DP3} with a learning-optimization hybrid strategy. This strategy allows the trained policy to fit new grippers without policy retraining or fine-tuning, as shown in Fig.\ref{fig:first-figure}\textcolor{blue}{b}. The policy first learns pick-and-place manipulation primitives following diffusion policy, where the training demonstration data is collected via a base gripper. During the policy's online inference, the base gripper is replaced with a new one. To ensure the generative trajectories are valid for new gripper configurations, we first incorporate the gripper-specific dimensional offsets into the policy inputs, and recast the traditional diffusion sampling process as a constrained-optimization process. The optimization introduces mathematical encoding of task and safety constraints (e.g.,  object grasping, avoiding collisions) to the denoising process, utilizing a quadratic programming optimization method to incrementally refine an initially noisy trajectory to conform to unseen gripper geometries. Crucially, this approach does not require policy retraining or fine-turning, thereby preserving flexibility for real-world deployment with minimal overhead.

We validate our approach with a Franka Panda manipulator equipped with various grippers for pick-and-place tasks—including 3D-printed fingertips of different heights, flexible fingertips, and the Robotiq 2F gripper—achieving robust performance without policy retraining. Key contributions are summarized as follows.
\begin{itemize}
    \item A learning-optimization strategy is introduced into the diffusion policy, dynamically optimizing the generation of trajectories to counteract the TCP offset problem caused by switching grippers, thus ensuring task completion. This ensures the trained policy can solve pick-and-place tasks with new grippers without retaining and fine-tuning operations.
    \item A general formulation of the task and safety constraints is designed for the optimization process. Moreover, a cumulative optimization is utilized to ensure the temporal consistency of the generated trajectory.
    \item Real-world experiments across different grippers show our method achieves 93.3\% and 70.0\% success rates for seen and unseen objects, respectively--outperforming baseline diffusion policies (23.3\% and 20.0\%).
\end{itemize}

\section{Related Work} 

Transferring robot manipulation knowledge between different embodiments can improve the policy's flexibility and generalizability for adapting to new tasks or hardware settings~\cite{wang2024scaling}, including the knowledge transitions between different grippers~\cite{zakka2022xirl} or robot manipulators~\cite{wang2024scaling,wang2024cross,tatiya2023transferring}. For example, Zakka et al.~\cite{zakka2022xirl} combine the temporal cycle-consistency method with imitation learning to learn an invariant feature space for different embodiments, empowering robot manipulators (target domain) to perform the tasks by imitating video demonstrations of human experts (source domain). While the feature correspondence between the source and target domains has to be re-trained if a new source or target domain is introduced. Wang et al.~\cite{wang2024cross} mitigate this limitation by learning a feature latent space for the source (a 7DoF manipulator) and target (a 6DoF manipulator) domains alongside the encoder and decoder for each domain. Then, they employ adversarial training and cycle consistency for latent alignment, which aligns the target encoder-decoder to that of the source domain, respectively. Similarly, in~\cite{tatiya2023transferring}, the authors employ latent alignment for source and target domains, achieving knowledge transferring between heterogeneous manipulators. By leveraging large-scale heterogeneous robot data, Wang et al.~\cite{wang2024scaling} propose a heterogeneous pre-trained transformer (HPT) that can perform different tasks across heterogeneous robot manipulators. 

However, the manipulation knowledge transition of these approaches relies on transfer learning or domain adaption/alignment~\cite{wang2024cross,wang2024scaling,tatiya2023transferring}, where introducing new hardware configuration (e.g., grippers) to the policy require fine-tuning policy~\cite{wang2024scaling} or retraining auxiliary networks (e.g., encoder and decoder)~\cite{wang2024cross,tatiya2023transferring}, limiting the flexibility of the policy. Our method circumvents these fine-tuning and retraining operations by integrating a learning-optimization hybrid strategy into diffusion policy~\cite{chi2023diffusion}, necessitating the introduction of task-oriented constraints for new hardware configurations and employing optimization during the policy online inference phase to ensure completion of the task.

Diffusion models have shown promise for solving decision-making tasks, including motion planning~\cite{10342382,mizuta2024cobl}, imitation learning~\cite{chi2023diffusion,Ze2024DP3,reuss2023goal}, and reinforcement learning~\cite{liang2023adaptdiffuser,venkatraman2024reasoning}. When solving decision-making tasks through generative models, the policy needs not only to accomplish the task goal but also to satisfy certain task constraints, such as collision avoidance. Some works add a residual loss to the training objectives if the task constraints are consistent during the policy training and inference~\cite{mizuta2024cobl,bastek2024physics}. A more flexible method is training diffusion models via classifier-free guidance\cite{ho2022classifier}, introducing conditioning variables that represent constraints into the policy training, such as physical constraints for guiding human motion generation~\cite{yuan2023physdiff}. However, the model conditioning encourages the generated samples to adhere to task constraints rather than strict guarantee constraints~\cite{christopher2024constrained}. Alternative post-processing methods draw constraints into the last denoising stage of the sample-generating process and obtain samples that satisfy the constraints by solving an optimization problem~\cite{maze2023diffusion,giannone2023aligning}. As the optimization problem does not consider the unknown data likelihood, post-processing may result in samples significantly deviating from the data distribution~\cite{romer2024diffusion}. To mitigate this issue, an iterative projection is integrated into the denoising process, confining each generated sample to the constraints. Some works employ it for sequential decision-making~\cite{christopher2024constrained}, which is time-consuming. Or model-based trajectory control~\cite{romer2024diffusion}, falling short of seamlessly adapting to a new model since introducing a new gripper changes the robot dynamics. Our approach utilizes projection alongside optimization techniques in the denoising process but mitigates the time consumption and makes the policy seamlessly adaptable to new grippers. 
\section{Background on diffusion policy} 
Denoising diffusion probabilistic models (DDPMs) \cite{ho2020denoising} are generative models that iteratively refine noisy data samples to match a target distribution.  They employ two complementary Markovian processes: a forward diffusion process that incrementally adds Gaussian noise to an original data sample $x^0$ over $k$ steps, and a reverse denoising process that removes the accumulated noise step by step. Building on this foundation, Diffusion Policy \cite{chi2023diffusion} adapts DDPMs for robotic manipulation. In this setting, a policy $\pi_{\theta}(A_t \mid S_t)$ is trained via behavior cloning to generate action sequences conditioned on visual and robot state observations. The model injects noise into the action samples during training and learns to predict this noise over multiple denoising steps, thereby guiding the policy to produce coherent actions at inference. Specifically, at test time, the reverse denoising process is iterated $K$ times to times to obtain a ``clean'' action sequence $A^0_t$. Concretely, assume a horizon $T_a$ and a state sequence $S_t = \{ s_{t - T_a + 1}, s_{t - T_a + 2}, \ldots, s_t \}$. The policy generates an action sequence $A_t = \{ a_t, a_{t+1}, \ldots, a_{t + T_a - 1} \}$ based on the learned diffusion mechanism. After executing $A_t$ on the robot, new sensor readings yield a subsequent state sequence $S_{t + T_a}$, from which the policy infers the next action sequence. By iterating this diffusion-based inference, the robot continuously adapts its actions to changing environment states, enabling robust long-horizon control.

\section{Methodology}
\subsection{Problem Statement}

We formalize the challenge of gripper-agnostic manipulation via diffusion policies through three core components:

\begin{itemize}
    \item \textbf{Gripper Configuration}: Let $\mathbb{G} \subset \mathbb{R}^{d_g}$ denote the space of two-finger gripper parameters encoding morphology, maximum width $w^{\max}$, and tool-center-point (TCP) $z^{\text{}}$, as shown in Fig. \ref{fig:grippers}. 
    \item \textbf{Observation Domain}: $\mathcal{O} = \mathcal{S}_{\text{sce}} \times \mathcal{S}_{\text{rob}}$ where $\mathcal{S}_{\text{sce}}$ represents scene observations (3D point clouds) and $\mathcal{S}_{\text{rob}} = SE(3) \times [0,g]$ the robot state (end-effector pose $\mathbf{x}_{\text{ee}} \in SE(3)$, gripper width $g$)\footnote{This paper forces on two-finger grippers problems in SE(3) space. SE(6)-space and three-finger grippers implementation is future work.}. $\mathbf{x}_{\text{ee}}$ is gripper-agnostic and reading from the robot. $g$ is specific for grippers.
    \item \textbf{Action Space}: $\mathcal{A} \subset \mathbb{R}^{d_a}$ containing end-effector displacements $\Delta\mathbf{x}_{\text{ee}}$ and gripper commands. 
\end{itemize}

The policy $\pi_\theta$ is trained on demonstrations $\mathcal{D} = \{\tau^{(i)}\}_{i=1}^N$ collected with a reference gripper $\mathbb{G}_0 \in \mathbb{G}$. Each trajectory $\tau = \{(\mathbf{o}_t, \mathbf{a}_t)\}_{t=0}^T$ satisfies: $\mathbf{o}_t = (\mathcal{S}^0_{\text{sce}}, \mathbf{x}_{\text{ee}}^0, g_t^0)$ and $\mathbf{a}_t \sim \pi_{\text{expert}}(\cdot|\mathbf{o}_t)$,
where superscript $0$ indicates $\mathbb{G}_0$ parameters.

During deployment with novel gripper $\mathbb{G}_i \neq \mathbb{G}_0$, the observation-action distribution shifts due to (1) visual/kinematic differences $\mathcal{O}_i \neq \mathcal{O}_0$, and (2) policy mismatch $p_{\theta}(\mathcal{A}|\mathcal{O}_i) \neq p_{\theta}(\mathcal{A}|\mathcal{O}_0)$. This manifests as trajectory divergence:
\begin{equation}
\|\tau_{1:T}^{\mathbb{G}_i} - \tau_{1:T}^{\mathbb{G}_0}\|_{\mathcal{W}} > \delta_{\text{tol}}\text{,}
\end{equation}
where $\mathcal{W}$ is the task-specific metric space and $\delta_{\text{tol}}$ the success threshold, e.g., objects cannot be grasped with shorter grippers, and collisions can result from using longer grippers.

\input{figure-exp-gripper}

To mitigate these issues, 
we develop policy $\pi_\theta^*$ that maintains task performance under gripper variation, combining (1) {gripper-invariant knowledge learning} and (2) {morphology-aware trajectory optimization} to achieve $\pi_\theta^*$ without policy retraining. The overview framework is shown in Fig. \ref{fig:framework}.

\input{figure-framework}
\input{figure-gmap}

\subsection{Learning Gripper-agnostic Grasping Knowledge}
Visuomotor policies, like Diffusion policy and 3D Diffusion Policy~\cite{chi2023diffusion,Ze2024DP3}, depend on visual observations to generate robot trajectories. However, swapping out different grippers during the online execution can alter visual observations (both RGB and point cloud inputs), as shown in Fig.\ref{fig:grippers}. Such alterations lead to out-of-distribution trajectory generations, reducing the task success rate\cite{liu2023towards}. To mitigate this limitation, we introduce a gripper-invariant \textit{grasping probability map} $\mathcal{G}_{\text{prob}}$ as an additional observation component, which captures object-centric grasp affordances  that are independent of end-effector geometry, thereby guiding the policy to focus on relevant object features rather than gripper-specific visual patterns. By decoupling object-related cues from the gripper's appearance, $\mathcal{G}_{\text{prob}}$ enhances the policy's robustness to variations in gripper morphology, maintaining stable task performance across different grippers.

We adopt the Generative Grasping CNN (GG-CNN) \cite{morrison2020learning} for $\mathcal{G}_{\text{prob}}$ synthesis from depth images. GG-CNN is pre-trained on the Cornell Grasping Dataset~\cite{lenz2015deep}, which contains 885 RGB-D images with annotated parallel-jaw grasps across 240 objects. However, real-world pick-and-place manipulations introduce two key challenges: (1) the hand-eye camera moving with the robot, causing scale variations in object pixels, and (2) lighting changes disturb depth sensor readings. These factors degrade GG-CNN's output stability, i.e., $\mathcal{G}_{\text{prob}}$, and destabilize policy training and inference performance. To address this issue, our solution involves: (1) threshold filtering: discard pixels with $\mathcal{G}_{\text{prob}} < 0.7$, (2) centroid computation: $\mathds{O} = \frac{1}{N}\sum_{i=1}^N (u_i,v_i)$ for remaining pixels, and (3) region masking: generate $\mathcal{G}^*_{\text{prob}}$ through circular masking ($r=30$ pixels) about $\mathds{O}$. The map $\mathcal{G}^*_{\text{prob}}$ satisfies:
\begin{equation}
    \mathcal{G}^*_{\text{prob}}(u,v) = \begin{cases}
        1 & \text{, if } \|(u,v) - \mathds{O}\|_2 \leq 30 \\
        0 & \text{, otherwise}
    \end{cases}
\end{equation}

This spatial filtering maintains grasp affordance information while eliminating outlier predictions caused by sensor noise, as shown in Fig.\ref{fig:gmap}. The policy observation consists of:
\begin{equation}
    \label{eq:obs}
    \mathcal{O}^* = \mathcal{G}^*_{\text{prob}}\times \mathcal{S}_{\text{sce}}\times \mathcal{S}_{\text{rot}},
\end{equation}

\subsection{Optimizing Trajectory Generation} 
\label{sec:optimization}
The training scheme of our policy $\pi_{\theta}$ is consistent with that of Diffusion Policy (DP)~\cite{chi2023diffusion}, i.e. DDPM, with observations $\mathcal{O}^*$ and MSE training loss. During inference, $\pi_{\theta}$ employs Denoising Diffusion Implicit Models (DDIM)\cite{song2021denoising} sampling with 10 denoising iterations. However, our policy introduces a strategy to enforce the generative trajectory to fit different grippers, comprising gripper-geometry aware mapping and safety-constrained trajectory  projection.

\textbf{Gripper mapping:} Gripper morphological variations induce the end-effector's pose discrepancies during identical object manipulation, primarily along: (1) vertical axis ($z$): tool-center-point offset, and (2) gripper state ($g$): grasping width differences. The discrepancies causes inconsistent actions predicted by $\pi_\theta$ across grippers.

Let $\mathbb{G}_{(0)}$ denote the reference gripper used during training, and $\mathbb{G}_{(i)}$ represent a novel gripper of category $i$. We define differentiable mapping functions $\mathds{M}_h(\cdot)$ and $\mathds{M}_g(\cdot)$ that project $\mathbb{G}_{(i)}$ parameters to the $\mathbb{G}_{(0)}$ basis: 
\begin{equation}
\label{eq:gripper_mapping}
\begin{aligned}
    z'_{(i)} &= \mathds{M}_h(z_{(i)}) = z_{(i)} + \Delta h_{(i)}, \\
    g'_{(i)} &= \mathds{M}_g(g_{(i)}) = \alpha_{(i)} g_{(i)},
\end{aligned}
\end{equation}
where $z_{(i)}$ is the measured height of end-effector equipped with $\mathbb{G}_{(i)}$, $\Delta h_{(i)} = z_{(0)} - z_{(i)}$ is the height offset from reference gripper, $g_{(i)} \in [g^{\min}, g^{\max}]$ is the real-time grasping width, $\alpha_{(i)} = {g^{\max}_{(0)}}/{g^{\max}_{(i)}}$ scales widths. The mapping parameters $\{\Delta h_{(i)}, \alpha_{(i)}\}$ are obtained through offline calibration with mechanical measurement of gripper dimensions. 

This transformation preserves the policy's internal representation while adapting to physical gripper properties, enabling zero-shot generalization to novel end-effectors. Current implementation focuses on translational pose adaptation, and the rotational compensation remains future work. During policy execution, the transformed pose $\mathcal{S}'_{\text{rob}} = (x,y,z'_{(i)},g'_{(i)})$ is fed to $\pi_\theta$ instead of $\mathcal{S}_{\text{rob}}$ in \eqref{eq:obs}, maintaining observation-space consistency across grippers.

\textbf{Safety-Constrained Trajectory Projection} 
While gripper mapping aligns geometric parameters, visual perception differences from gripper morphology can still induce unsafe trajectory variations. To guarantee constraint satisfaction, we integrate a projection layer into the DDIM denoising process~\cite{song2021denoising}. The modified reverse diffusion step becomes:
\begin{equation}\label{eq:modified_ddim}
  \mathbf{a}^{k-1}_{t} = \text{Proj}_{\mathcal{C}}\left(\mu_{k}(\mathbf{a}^k_{t}, \epsilon_{\theta}(\mathbf{a}^k_{t},\mathbf{o}_{t},k))\right),
\end{equation}
where $\text{Proj}_{\mathcal{C}}(\cdot)$ enforces safety constraints $\mathcal{C}$ through the following two steps.

\noindent\paragraph{\textbf{Constraint-aware denoising}}
For efficiency, projection activates only in the final denoising steps ($k \leq 2$). At each step $k$, we solve a quadratic program problem:

\begin{equation}\label{eq:safety_optimization}
\begin{aligned}
    \nu^{k*}_t &= \underset{\nu^k_t}{\arg\min} \|\nu^k_t\|^2_2 \\
    \text{s.t. } & \mathcal{S}'_{\text{rob}}(z)_t + \Phi\left(\mathbf{a}^{k}_t\right) + \nu^k_t \geq \epsilon_{\text{safe}}
\end{aligned}
\end{equation}
where $\Phi(\cdot)$ maps latent actions to Cartesian displacement, which is denormalization in our case, $\epsilon_{\text{safe}} = 0.01$ m (safety margin), and $\nu^k_t$ is the minimal corrective offset.

\noindent\paragraph{\textbf{Temporal consistency enforcement}} 
To maintain safety over the policy's $T_a$-step action horizon ($j \in [0,T_a-1]$), we extend \eqref{eq:safety_optimization} with cumulative constraints:
\begin{equation}\label{eq:sequence_constraint}
\mathcal{S}'_{\text{rob}}(z)_{t} + \sum_{r=0}^{j}\Phi(\mathbf{a}^{k}_{t+r}) + \nu^k_{t+j} \geq \epsilon_{\text{safe}}
\end{equation}

The projected actions $\mathbf{a}^{k*}_t = \Phi^{-}[\Phi(\mathbf{a}^{k}_t) + \nu^{k*}_t]$ guarantee:
\begin{equation}
\mathbb{P}\left(\bigcap_{j=0}^{T_a} \{\mathcal{S}'_{\text{rob}}(z)_{t+j} \geq \epsilon_{\text{safe}}\}\right) = 1,
\end{equation} 
indicating the cumulative trajectory is always safe, with an example of $\mathcal{S}'_{\text{rob}}(z)_{t+1}=\mathcal{S}'_{\text{rob}}(z)_t+\Phi(\mathbf{a}^{k*}_t)$.

\begin{algorithm}[!t]
    \caption{Gripper-Aware Trajectory Generation}\label{alg:safe_diffusion}
    \setstretch{1}
    \begin{algorithmic}[1]
    
        \Require Novel gripper $\mathbb{G}_i$, Observation $\mathbf{o}_t$, Safety margin $\epsilon_{\text{safe}}$, trained noise predicted $\epsilon_\theta$
        \Ensure Safe trajectory $\tau = \{\mathbf{a}_{t:t+T_a-1}\}$
        
        \noindent \textcolor{gray}{// \textbf{Online Inference:}}
        \State $\mathcal{S}'_{\text{rob}} \leftarrow \mathds{M}_h(z_{(i)}) = z_{(i)} + \Delta h_{(i)}$ \textcolor{gray}{// \textbf{Gripper Mapping:}}
        \State $g'_{(i)} \leftarrow \mathds{M}_g(g_{(i)}) = \alpha_{(i)} g_{(i)}$
        \State $\mathbf{\tilde{o}}_t \leftarrow (\mathcal{G}^*_{\text{prob}}, \mathcal{S}^{ }_{\text{sce}}, \mathcal{S}'_{\text{rob}})$
    
        \State $\mathbf{a}^{K}_t \sim \mathcal{N}(0, \mathbf{I})$ \textcolor{gray}{// \textbf{Diffusion Process:}}
        \Repeat
            \State $k \leftarrow K-1$, and $K \leftarrow K-1$
            \State $\mathbf{a}^{k}_t \leftarrow \mathcal{N}\big(\mu_{k}(\mathbf{a}^{k+1}_{t}, \epsilon_{\theta}(\mathbf{a}^{k+1}_{t},\mathbf{\tilde{o}}_t,k+1)), 0\big)$
            \State \textbf{if} $k \leq 1$: \textcolor{gray}{// \textbf{Safety Projection:}}
            \State\quad\ \textbf{for} $j \leftarrow 0$ \textbf{to} $T_a-1$ \textbf{do}
            \State\quad\quad\ \ $\nu^{k*}_{t+j} \leftarrow \arg\min\limits_{\nu} \|\nu^k_{t+j}\|^2_2$
            \State\quad\quad\ \ $\text{s.t. } \mathcal{S}'_{\text{rob}}(z)_{t}+\sum\limits_{r=0}^{j}\Phi(\mathbf{a}^{k}_{t+r}) + \nu^k_{t+j} \geq \epsilon_{\text{safe}}$
            \State\quad\ $\mathbf{a}^{k}_{t:t+T_a-1} \leftarrow \Phi^{-}[\Phi(\mathbf{a}^{k}_{t:t+T_a-1}) + \nu^{k*}_{t:t+T_a-1}]$
        \Until{$K=0$}
        \State $\tau \leftarrow \text{Decode}(\mathbf{a}^{0}_{t:t+T_a-1})$
    \end{algorithmic}
\end{algorithm} 

\section{Experiments} 
To evaluate our proposed method, we designed experiments to (i) evaluate the stability of grasping probability maps under different object and robot pose conditions, (ii) validate different policy's generalization for pick-and-place tasks across diverse two-finger grippers without retraining the policy, and (iii) assess the generated trajectory's safety and task completion issues across different policies.

\subsection{Experiment Setup}
The experimental setup employs a Franka Emika Panda robotic arm equipped with interchangeable end-effectors, including a native parallel gripper augmented with custom 3D-printed fingertips and a Robotiq 2F-85 adaptive gripper, as shown in Fig. \ref{fig:grippers}. Real-time control is managed via a workstation running Ubuntu 20.04 with a real-time kernel to ensure deterministic low-latency operation, while Diffusion Policy framework is trained and executed on another workstation with a NVIDIA RTX-4090 GPU and an AMD 7950x CPU. Perception integrates an egocentric Azure Kinect DK camera and a fixed Intel RealSense D455 camera. Sensor data synchronizations are facilitated through ROS noetic.

\subsection{Data Collection and Training}
\textit{1) Data Collection:} Our dataset comprises 59 expert demonstration trajectories of pick-and-place tasks with the gripper $\mathbb{G}_0$ in Fig. \ref{fig:grippers}. Multimodal observations are captured through three synchronized modalities: (1) Proprioceptive states: including the 6-DoF end-effector pose and continuous gripper width measurements obtained via the Franky API\footnote{\href{https://github.com/TimSchneider42/franky}{https://github.com/TimSchneider42/franky}}; (2) 3D point clouds are acquired via the D455 camera and are processed with farthest point sampling. To estimate the gripper-agnostic \textit{grasping probability map} $\mathcal{G}^*_{\text{prob}}$, we adapt the grasping probability representation from \cite{10582538} to process the Kinect camera's depth and RGB images; (3) Kinesthetic demonstration signals: recorded via a joystick controller operating at 2Hz, encoding relative Cartesian displacements alongside a one-hot encoded gripper action (open/closed).

\textit{2) Training:}  We select the Simple DP3 implementation \cite{Ze2024DP3} as our policy's backbone for efficiency. To integrate the \textit{grasping probability map} $\mathcal{G}^*_{\text{prob}}$ into the policy framework, we adapt Simple DP3 to Simple DP3 MM (Multi-Modal Simple Diffusion Policy 3D). This architecture concatenates three encoded inputs: Proprioceptive states and point cloud features processed identically to Simple DP3; $\mathcal{G}^*_{\text{prob}}$ encoded through a ResNet-18 network. The model retains the same pruned UNet backbone as Simple DP3. Training follows the original Simple DP3 procedure but extends the duration to 8,000 epochs to accommodate the additional modality.

\def\figureww{2cm}
\def\figurehh{3cm}
\begin{figure}[!t]
  \centering
  \captionsetup{singlelinecheck=off}
    \begin{tikzpicture}
        \node(border)[minimum height=3cm, minimum width=8.8cm, draw=none, dashed] at (0,0){};
        \node[inner sep=0pt,anchor=north east](gmap1) at ($(border.north east)+(0,0)$){\includegraphics[height=\figurehh]{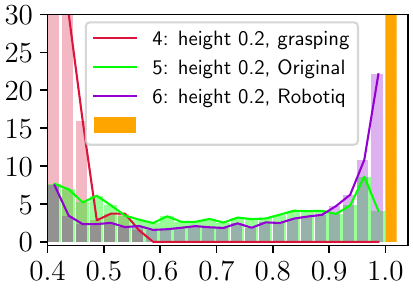}};
        \node[inner sep=0pt,anchor=east](gmap2) at ($(gmap1.west)+(0,0)$){\includegraphics[height=\figurehh]{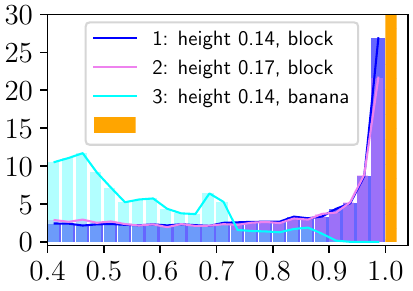}};
        \node[inner sep=0pt,black,rotate=90,anchor=center] at ($(gmap2.west)+(-0.13,0)$){\footnotesize percentage (\%)};
        \node(gtext)[inner sep=0pt,black,anchor=north] at ($(gmap2.south east)+(0,0.05)$){\footnotesize grasping probabilities ($>$ 0.4)};
        \node[inner sep=0pt,red,anchor=center] at ($(gmap2.east)+(-2.35,0.16)$){\footnotesize $\mathcal{G}^*_{\text{prob}}$};
        \node[inner sep=0pt,red,anchor=center] at ($(gmap1.east)+(-2.35,0.16)$){\footnotesize $\mathcal{G}^*_{\text{prob}}$};
        \node[inner sep=1.5pt,anchor=north] at (gmap2.south|-gtext.south) {\footnotesize (a) different heights, objects};
        \node[inner sep=1.5pt,anchor=north] at (gmap1.south|-gtext.south) {\footnotesize (b) different heights, grippers};
    \end{tikzpicture}
  \caption{Grasping probabilities map under different conditions.}
    \label{fig:exp-gmap}
\end{figure}

\begin{table}[!t]
  \begin{center}
  \renewcommand{\arraystretch}{1.2}
  \small
    \caption{KL divergence between distributions in Fig. \ref{fig:exp-gmap}.}
    \label{tab:kl}
    \resizebox{0.95\columnwidth}{!}{
    \begin{tabular}{l|c|c|c|c|c} 
        \toprule[1pt]\midrule[0.2pt]
       & {$\mathcal{G}^{(2)}_{\text{prob}}$} & $\mathcal{G}^{(3)}_{\text{prob}}$ & $\mathcal{G}^{(4)}_{\text{prob}}$ & $\mathcal{G}^{(5)}_{\text{prob}}$ & $\mathcal{G}^{(6)}_{\text{prob}}$\\
      \midrule
      KL($\mathcal{G}^{(i)}_{\text{prob}}$,$\mathcal{G}^{(1)}_{\text{prob}}$) & 0.009 & 0.929 & 0.276 & 2.271 & 0.052\\
    \midrule[0.2pt]
    KL(\textcolor{red}{$\mathcal{G}^{*(i)}_{\text{prob}}$},\textcolor{red}{$\mathcal{G}^{*(1)}_{\text{prob}}$}) & 0.000 & 0.000 & 0.000 & 0.000 & 0.000\\
    \midrule[0.2pt]\bottomrule[1pt]
    \end{tabular}
    }
  \end{center}
\end{table} 

\subsection{Real-world Experiment}
\textbf{1) Evaluation of Grasping Probability Map}: 
We evaluate the stability of grasping probability maps under three perturbation scenarios, including graspable object variation, gripper morphology, and viewpoint changes (i.g. end-effector heights). We define stability as the extent to which the probability distribution of these maps remains consistent across the different scenarios. If the variance in distribution is large, the resulting shift can adversely affect the policy’s trajectory generation from training to inference. Fig.~\ref{fig:gmap} visualizes raw GG-CNN outputs $\mathcal{G}_{\text{prob}}$ versus our processed maps $\mathcal{G}^*_{\text{prob}}$ across these conditions. Fig. \ref{fig:exp-gmap} presents the difference between $\mathcal{G}_{\text{prob}}$ and $\mathcal{G}^*_{\text{prob}}$ through normalized probability histograms, here only shows high-probability parts ($>$ 0.4). The baseline $\mathcal{G}_{\text{prob}}$ demonstrates multimodal distributions versus our unimodal concentration. Moreover, Tab. \ref{tab:kl} quantifies the distribution difference across conditions via KL divergence, indicating $\mathcal{G}^*_{\text{prob}}$ provides stable grasping information against perturbations. The stabilized grasp cues enable the policy to generate spatially consistent approach trajectories despite sensory perturbations, which is critical for reliable manipulation across grippers. 

\textbf{2) Policy Performance Comparison across Grippers}:

(a) \textbf{Setup}: Our policy and baselines are evaluated in different pick-and-place task setting, including utilizing different grippers and manipulating different objects. Notably, all training demonstrations included only one type of gripper ($\mathbb{G}_0$) and graspable object (a pink block).

(b) \textbf{Success Rate Definition}: Picking the object from the tabletop and placing it on a box is a successful trial. All other outcomes are deemed failures, including failure to grasp the object or dropping it during manipulation.

(c) \textbf{Baseline and Ablation Study Setting}: 
\begin{itemize}
  \item Diffusion Policy (DP)~\cite{chi2023diffusion}:  Implements a baseline framework for temporal action sequence generation through diffusion modeling. DP receives two viewpoint RGB images from the Realsense and Kinect cameras.
  \item Diffusion Policy 3D (DP 3D)~\cite{Ze2024DP3}: Extends the Diffusion Policy paradigm by incorporating 3D spatial reasoning. It receives point clouds from the Realsense camera as visual observations.
  \item Ours w/o Projection: The trajectory optimization strategy in \ref{sec:optimization} is not activated in the online inference phase, including gripper mapping and safety-constrained trajectory projection.
  \item Ours w/o $\mathcal{G}^*_{\text{prob}}$: Utilizing RGB images from the Kinect camera to train policy, instead of gripper-agnostic grasping knowledge $\mathcal{G}^*_{\text{prob}}$.
\end{itemize}

\begin{table}[!t]
  \begin{center}
    \caption{Pick-and-place task across grippers.}
    \label{tab:comparison}
    \resizebox{\columnwidth}{!}{
    \begin{threeparttable}
        \begin{tabular}{l|c|c|c|c|c|c|c} 
            \toprule[1.5pt]
            \midrule[0.2pt]
            \multicolumn{8}{c}{\textcolor{Salmon}{Seen object:} {Block} (unit: \%)}\\
            \midrule
           Method & $\mathbb{G}_0$ & $\mathbb{G}_1$ & $\mathbb{G}_2$ & $\mathbb{G}_3$ & $\mathbb{G}_4$ & $\mathbb{G}_5$ & $\mu$ \\
          \midrule
          Diffusion Policy & 20.0 & 0.0 & 60.0 & 40.0 & 0 & 40.0 & 26.7 \\
          Diffusion Policy 3D & 20.0 & 0.0 & 60.0 & 60.0 & 0.0 & 0.0 & 23.3\\
          DP + Projection & 100.0 & - & - & - & - & - & -\\
          DP 3D + Projection & 80.0 & - & - & - & - & - & -\\
          
          Ours \textcolor{red}{w/o} Projection & 100.0 & 0.0 & 60.0 & 40.0 & 0.0 & 0.0 & 33.3 \\
          Ours \textcolor{red}{w/o} $\mathcal{G}^*_{\text{prob}}$ & 80.0 & 20.0 & 40.0 & 80.0 & 100.0 & 0.0 & 53.3 \\
          Ours & \textbf{100.0} & \textbf{80.0} & \textbf{100.0} & \textbf{80.0} & \textbf{100.0} & \textbf{100.0} & \textbf{93.3}\\

        \midrule[0.2pt]
        
        \multicolumn{8}{c}{\textcolor{Salmon}{Unseen object:} {Banana} (unit: \%)}\\
        \midrule
          Diffusion Policy & 20.0 & 0.0 & 40.0 & \textbf{60.0} & 40.0 & 20.0 & 30.0\\
          Diffusion Policy 3D & 20.0 & 0.0 & 40.0 & 40.0 & 20.0 & 0.0 & 20.0\\
          Ours \textcolor{red}{w/o} Projection & 80.0 & 0.0 & 40.0 & 40.0 & 0.0 & 20.0 & 30.0 \\
          Ours \textcolor{red}{w/o} $\mathcal{G}^*_{\text{prob}}$ & 60.0 & 0.0 & 40.0 & \textbf{60.0} & 40.0 & 0.0 & 33.3 \\
          Ours & \textbf{80.0} & \textbf{60.0} & \textbf{100.0} & \textbf{60.0} & \textbf{60.0} & \textbf{60.0} & \textbf{70.0}\\
        \midrule[0.2pt]\bottomrule[1.5pt]
        \end{tabular}
    
        \begin{tablenotes}[para,flushleft]
            \small
            - The training data is collected with the base gripper \textcolor{red}{$\mathbb{G}_0$} and a pink block. \\
            - The initial pose and position of the object are identical across tests.\\
            - \textcolor{red}{$\mathbb{G}_i$} indicates different grippers reported in Fig. \ref{fig:grippers}. \textcolor{red}{w/o}: without this module.
            - Every method is validated 5 times for each gripper, totaling 30 evaluations for calculating the average success rate \textcolor{red}{$\mu$} across different grippers.\\
            
            
        \end{tablenotes}
    \end{threeparttable}
    }
  \end{center}
  \vspace{-2em}
\end{table} 

(e) \textbf{Quantitative results}:
Our method demonstrates generalization across diverse gripper geometries relative to baselines. As shown in Table~\ref{tab:comparison}, our method achieves an average success rate of \textbf{93.3\%} on all gripper configurations for the seen graspable object (block). In contrast, for the original gripper \(\mathbb{G}_0\), both DP and DP 3D exhibit notably lower success rates (e.g., DP succeeds only 20\%, while DP 3D fails entirely), primarily due to their imprecise motion generation, resulting in collisions with the table. When transitioning to shorter grippers (\(\mathbb{G}_2\) and \(\mathbb{G}_3\)), DP and DP 3D can complete the task, as the reduced gripper length diminishes the likelihood of collisions. Notably, our policy without the Projection (``Ours w/o Projection'') is capable of successfully completing the task for the original gripper \(\mathbb{G}_0\); however, it fails for grippers that deviate more significantly from \(\mathbb{G}_0\) (specifically, \(\mathbb{G}_1\), \(\mathbb{G}_4\), and \(\mathbb{G}_5\)). Furthermore, for the ablation setting with ``Ours w/o $\mathcal{G}^*_{\text{prob}}$'', the performance on \(\mathbb{G}_0\), \(\mathbb{G}_3\), and \(\mathbb{G}_4\) is comparable to, or slightly worse than, that of our full policy. However, when evaluated on grippers that appear different from \(\mathbb{G}_0\) from the perspective of the Kinect camera (namely, \(\mathbb{G}_1\) and \(\mathbb{G}_5\)), its performance degrades (the success rate ranging from 0 to 20\%). These results underscore the critical roles of both the Projection and gripper-agnostic grasping knowledge \(\mathcal{G}^*_{\text{prob}}\) in ensuring robust generalization across varying gripper geometries. Our method demonstrates robust generalization to unseen object geometries compared to baseline approaches, as evidenced by experiments with objects such as bananas. Our method achieves an average success rate of 70\%, outperforming baseline policies. While DP and DP 3D occasionally succeed (30\% and 20\%, respectively) when the unseen object is placed identically to training demonstrations, their reliance on position-specific motion primitives limits adaptability. For instance, DP's rigid trajectory generation fails to adjust to novel object shapes, often resulting in misaligned grasps (e.g., sliding off the curved surface of a banana). 

\input{figure-exp-safety}

\input{figure-exp-horizon}

When object positions are randomized, both DP and DP 3D collapse entirely (success rate is 0). Notably, our projection module plays a pivotal role in handling unseen geometries. Compared to the ablation without gripper-agnostic grasping knowledge ``Ours w/o \( \mathcal{G}^*_{\text{prob}} \)'', our policy exhibits a better understanding of grasping unseen objects by leveraging gripper-agnostic grasping knowledge. Similarly, our policy without the Projection module (``Ours w/o Projection'') is still able to locate the grasping point. However, it fails for grippers that deviate more significantly from \( \mathbb{G}_0 \) (specifically, \( \mathbb{G}_2 \), \( \mathbb{G}_4 \), and \( \mathbb{G}_5 \)). These results highlight the limitations of baseline methods in disentangling object geometry from positional priors. This capability is critical for real-world applications where object and experiment setup diversity are inherent.

\textbf{3) Safety Analysis}:
We analyze how \textbf{gripper mapping} and \textbf{safety-constrained trajectory projection} ensure safe trajectory generation and successful task completion across different grippers. As shown in Fig.~\ref{fig:safety}, for Gripper \( \mathbb{G}_0 \), as mentioned in quantitative results, DP and DP 3D often result in collisions. After introducing the \textbf{safety-constrained trajectory projection} to DP and DP 3D on the inference phase, the generated motions are projected within the defined safety constraints, enabling both policies to successfully complete the task without collisions, as shown in Fig. \ref{fig:safe-a}. When testing on gripper \( \mathbb{G}_1 \), which is longer than gripper \( \mathbb{G}_0 \) with TCP's shift +3.5cm, as shown in Fig. \ref{fig:safe-b}, DP, DP 3D, and our policy without the projection module consistently lead to table collisions. In contrast, our policy ensures collision-free execution by correctly adjusting the motion to the extended gripper length. For the shorter gripper \( \mathbb{G}_4 \), our policy without \textbf{gripper mapping} fails to adapt the motion to the shorter gripper compared to \( \mathbb{G}_0 \), where TCP's shift -4cm, resulting in a grasping point that is too high above the object. However, our full policy successfully generates adaptive motions that account for the shorter gripper length, ensuring accurate and stable object grasping. Fig. \ref{fig:real-world-horizon} displays our method's rollout of pick-and-place tasks with different grippers.

\section{CONCLUSIONS AND OUTLOOK}

This paper proposes a diffusion-based policy for transferring pick-and-place knowledge across different grippers. This knowledge transition does not require retraining or fine-turning the policy with the new gripper's configuration. Instead, it only needs to introduce the configuration in the policy inference phase and make the generated trajectories satisfy safety constraints, ensuring the successful completion of pick-and-place tasks. To enhance the flexibility of integrating different grippers into our method, the gripper's configurations can be described with free-form language instructions during the policy inference phase.

\normalem
\bibliographystyle{./IEEEtran} 
\bibliography{./IEEEexample}

\end{document}